\documentclass[conference]{IEEEtran}
\IEEEoverridecommandlockouts
\usepackage{booktabs, colortbl, xcolor, pifont, relsize}
\newcommand{\cmark}{\ding{51}} % 定义对号
\newcommand{\xmark}{\ding{55}} % 定义错号

\usepackage{cite}
\usepackage{amsmath,amssymb,amsfonts}
\usepackage{algorithmic}
\usepackage{graphicx}
\usepackage{textcomp}
\usepackage[table]{xcolor}
\usepackage{booktabs}
\usepackage{multirow}
\usepackage{newtxtext} 
\usepackage{fancyhdr}
\definecolor{correct}{rgb}{0.0, 0.5, 0.0}  % 绿色
\definecolor{wrong}{rgb}{0.8, 0.0, 0.0}   % 红色
\def\BibTeX{{\rm B\kern-.05em{\sc i\kern-.025em b}\kern-.08em
    T\kern-.1667em\lower.7ex\hbox{E}\kern-.125emX}}
\begin{document}
\pagestyle{fancy}        % use fancyhdr for all pages
\fancyhf{}               % clear header and footer
\fancyfoot[C]{\thepage}  % page number in the centre of the footer
\renewcommand{\headrulewidth}{0pt}
\renewcommand{\footrulewidth}{0pt}
\thispagestyle{fancy}    % also apply to the first page

\def\method{\texttt{Dyna-Pruner}}
\title{DYNA-PRUNER: Input-Adaptive Data–Model Co-Pruning for Efficient and Scalable Spatio-Temporal Media Prediction}

\author{
    \IEEEauthorblockN{
        Fuyan Zhang$^{1}$, Yuqi Li$^{2}$, Qing Xu$^{3}$, Yingli Tian$^{2}$, Edmond S.L. Ho$^{4, *}$%
        \thanks{$^{*}$Corresponding author.}
    }
    \vspace{0.15cm} 
    \IEEEauthorblockA{
        \textit{$^1$University of Edinburgh} \\
        \textit{$^2$The City College of New York, City University of New York} \\
        \textit{$^3$Institute of Computing Technology, Chinese Academy of Sciences} \\
        \textit{$^4$University of Glasgow}
    }
}

\maketitle

\begin{abstract}

Spatio-temporal prediction supports radar/satellite nowcasting and city-scale traffic monitoring, but modern models are often too expensive for real-time deployment. This stems from a mismatch between dense computation and strong input-dependent redundancy (e.g., calm seas or clear skies). To enable automated, resource-aware architecture optimization in scalable media analysis, we propose \method{}, an end-to-end framework for \textbf{input-dependent} co-pruning of data and model structure. A \textbf{shared-importance synchronization mechanism} generates coupled masks that prune redundant regions and their corresponding computational units (e.g., convolutional filters), yielding per-sample sparse sub-networks at inference time. Experiments on \textbf{WeatherBench, SEVIR, and TaxiBJ} show seamless integration with \textbf{CNN, RNN, and Transformer} backbones, reducing FLOPs by up to 70\% and achieving a \textbf{2.5$\times$} speedup on NVIDIA Jetson AGX Orin with negligible accuracy loss ($<1\%$).

\end{abstract}
\begin{IEEEkeywords}
Spatio-temporal Prediction, Collaborative Pruning, Data Pruning, Model Compression, Lightweight Models.
\end{IEEEkeywords}

\section{Introduction}
\label{sec:introduction}

%Spatio-temporal prediction is foundational to critical domains such as meteorological forecasting, oceanography, and urban traffic management~\cite{bi2023accurate, wu2024earthfarsser}. While deep neural networks have achieved strong accuracy by capturing complex dependencies, their deployment is often bottlenecked by computational overhead. This inefficiency stems from a fundamental mismatch: {standard architectures employ a ``dense'' computation paradigm that treats all inputs equally, whereas real-world spatio-temporal data is characterized by extreme sparsity and dynamic redundancy}~\cite{wu2023pastnet, gao2021phygeonet, yu2018spatio, raaisaanen2007reliable, mohan2020spatio}. For instance, in weather forecasting, vast regions of ``calm seas'' or ``clear skies'' often contain negligible dynamic information compared to localized storm cells, yet existing CNNs~\cite{gao2022simvp, he2022convolutional}, RNNs~\cite{wang2022predrnn, cho2014learning}, and Transformers~\cite{dosovitskiy2021an} uniformly process every pixel, wasting significant energy on low-information regions. This dense inference paradigm restricts the deployment of advanced models on resource-constrained edge devices and IoT sensors, where real-time, on-site insights are most critical.

Spatio-temporal prediction is foundational to a wide range of media- and sensing-driven applications such as meteorological forecasting, oceanography, and urban traffic management~\cite{wu2024earthfarsser}. In these settings, the inputs are often video-like spatio-temporal tensors (e.g., radar/satellite imagery sequences or citywide mobility maps), where accurate and timely forecasts are essential for downstream decision-making. While deep neural networks have achieved unprecedented accuracy by capturing complex dependencies, their deployment remains bottlenecked by massive computational overhead. This inefficiency stems from a fundamental mismatch: standard architectures employ a “dense” computation paradigm that treats all inputs equally, whereas real-world spatio-temporal data is characterized by extreme sparsity and dynamic redundancy~\cite{wu2024pastnet, gao2021phygeonet, yu2018spatio, mohan2020spatio,li2026comprehensive,li2025preference}. For instance, in weather forecasting, vast regions of ``calm seas'' or ``clear skies'' often contain negligible dynamic information compared to localized storm cells, yet existing CNNs ~\cite{gao2022simvp, he2022convolutional}, RNNs~\cite{wang2022predrnn}, and Transformers~\cite{dosovitskiy2021an} uniformly process every pixel, wasting significant energy on low-information regions. This rigid paradigm severely restricts the deployment of advanced models on resource-constrained edge devices and IoT sensors, where real-time, on-site insights are most critical.

To improve the trade-off between accuracy and efficiency in scalable media analysis, automated, resource-aware architecture optimization (e.g., neural architecture search) and model compression are increasingly important, yet most deployed networks still execute a fixed computational graph and cannot adapt cost to the {instantaneous information density} of each input. Existing model compression techniques, such as weight pruning~\cite{li2023model, choudhary2020comprehensive, huang2024magicfight, huang2025dive, DBLP:conf/mm/ShiZWZZL24}, primarily target internal parameter redundancy but are largely {input-agnostic} and fail to adapt the model's computational graph to the input. Conversely, standalone data simplification or sparse sampling~\cite{bazjanac2007reduction} focuses on input reduction but ignores the potential for model-side optimization. {Crucially, both paradigms lack a dynamic link between the input's informational content and the model's active computational path.} This gap motivates a central question: \textit{Can we learn a compression policy end-to-end that jointly optimizes both data and model sparsity on a per-input basis?}

In this paper, we propose \textbf{Dyna-Pruner}, a novel framework for {dynamic, collaborative data and model pruning}. Unlike static pruning, the ``dynamic'' nature of our framework refers to its {on-the-fly adaptation}: it enables the model to learn not only \textit{how} to predict but also \textit{where} to look and \textit{what} resources to activate for each specific sample, effectively realizing an input-conditioned sparse sub-network at inference time. At the heart of Dyna-Pruner is a {shared-importance synchronization mechanism} that generates two interconnected, learnable masks: a {Data Mask}~\cite{he2022masked} that adaptively nullifies redundant input regions, and a \textbf{Model Mask} that synchronously prunes corresponding computational units (e.g., convolutional filters) via receptive-field mapping~\cite{he2022convolutional}. By explicitly coupling these two dimensions, Dyna-Pruner forces the model to concentrate its limited computational budget exclusively on high-value, physically active regions.

Our main contributions are threefold: 
\begin{itemize}
    \item We propose the first {dynamic collaborative pruning} framework that \emph{unifies} data and model sparsification into a single end-to-end optimizable objective, moving beyond traditional input-agnostic compression.
    \item We introduce a {shared-importance synchronization} mechanism, where a continuous importance field $S$ enables \textit{per-input} adaptation through receptive-field aggregation and Straight-Through Estimator (STE) based training.
    \item Extensive evaluations on three diverse spatio-temporal benchmarks demonstrate that Dyna-Pruner significantly reduces computational complexity, achieving up to a {70\% reduction in FLOPs} and {2.5$\times$ practical speedup} on edge hardware with negligible accuracy degradation.
\end{itemize}

\section{Related Work}

\paragraph{Spatio-Temporal Prediction.} High-performance backbones, including CNNs~\cite{gao2022simvp}, RNNs~\cite{wang2022predrnn}, and Transformers~\cite{gao2022earthformer,li2025frequency}, have pushed the boundaries of accuracy in weather and traffic forecasting. Foundation models like Triton~\cite{wu2025triton} further enhance global modeling but incur prohibitive computational costs~\cite{bi2023accurate, gao2025oneforecast}. Our work addresses this bottleneck by introducing a lightweight computation paradigm for these heavy backbones.

\paragraph{Static vs. Dynamic Pruning.} Traditional structured pruning~\cite{li2023model, kumar2021pruning, li2023less, li2024pruning, li2026sepprune} primarily targets parameter redundancy but is typically \textit{static} and \textit{input-agnostic}. While data simplification~\cite{bazjanac2007reduction} reduces input size, it rarely co-optimizes with the model's architecture. \method{} differs by establishing a dynamic, sample-dependent link between input information density and the model's active computational path.

\paragraph{Dynamic Sparse Training.} Emerging sparse training methods like DynST~\cite{wu2025dynst} and automated graph lottery tickets~\cite{zhang2024graph, zhang2024heads} adapt to resource constraints. However, they focus primarily on model-side sparsity. Our framework introduces a \textit{shared-importance field} to synchronize data and model pruning, specifically targeting the unique redundancy in spatio-temporal physical signals.

\paragraph{Neural Operators and Physics.} Physics-informed models, such as PINNs~\cite{raissi2019physics} and Neural Operators~\cite{li2021fourier, raonic2024convolutional}, learn mappings for solving complex PDEs~\cite{hao2024dpot}. We complement this field by providing an efficient inference path that concentrates resources on physically active regions, maintaining high fidelity in critical dynamic areas.

\section{Method}
\subsection{Framework Overview}
\label{subsec:framework_overview}

\begin{figure*}[htbp]
\centering
\includegraphics[width=0.9\textwidth]{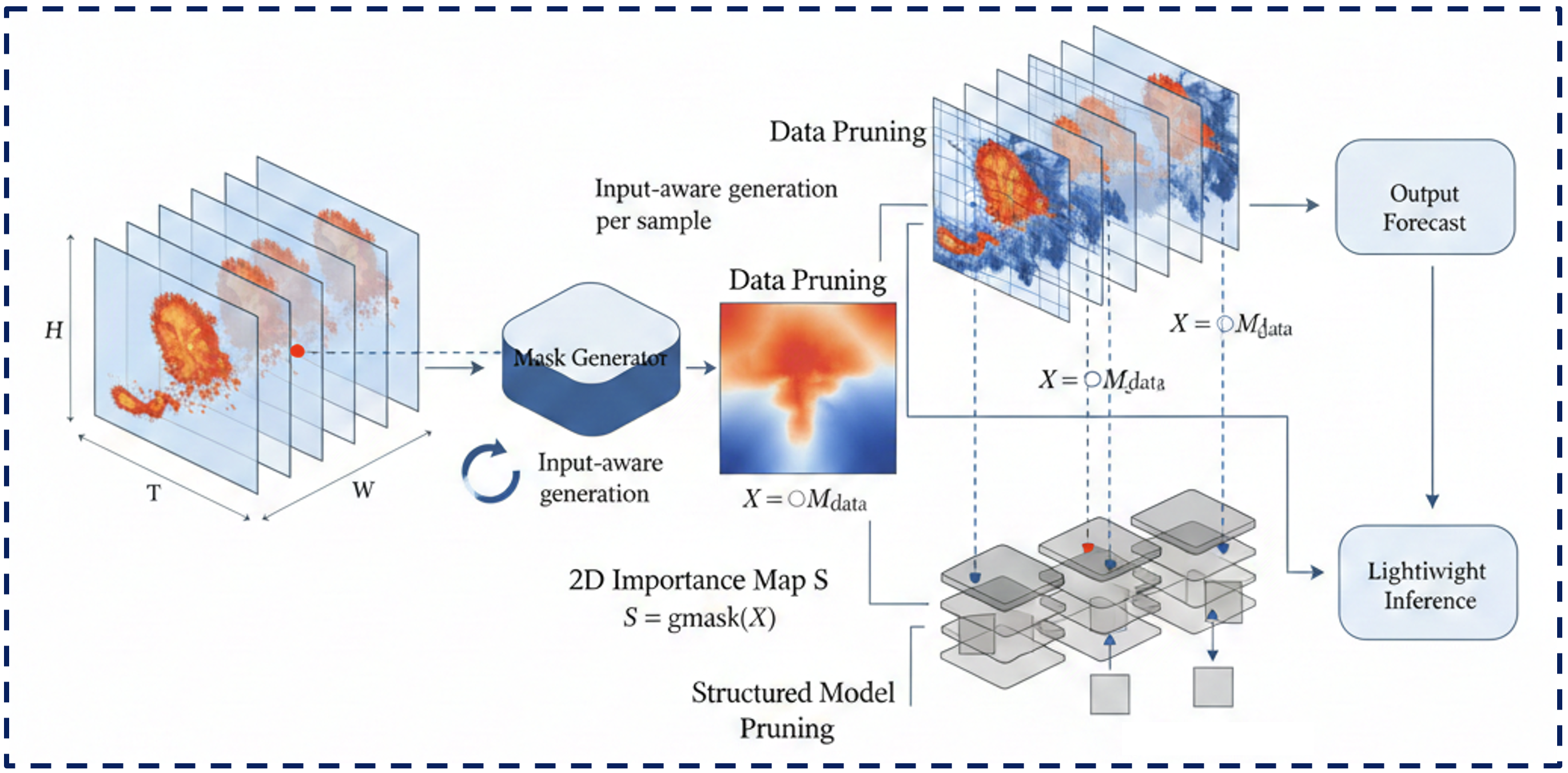}
\caption{An overview of the \method{} framework. This diagram illustrates how the collaborative pruning module dynamically generates a Data Mask and a Model Mask from the input data. These masks guide a sparse model to process data efficiently, and the entire system is optimized end-to-end.}
\label{fig:framework}
\end{figure*}

An overview of \method{} is illustrated in Figure~\ref{fig:framework}. Our core idea is to replace the traditional dense computation paradigm with one where the model's computational load adapts to the input's information density. At the heart of \method{} is a \textbf{collaborative pruning module} that dynamically generates two interconnected masks from dense spatio-temporal data: a \textbf{Data Mask} ($M_{\text{data}}$) and a \textbf{Model Mask} ($M_{\text{weight}}$). The data mask identifies and filters redundant input regions to form a sparse input, while the model mask synchronously prunes the corresponding computational units, creating a sparse computational path. This sparse model then processes the sparse data in a \textbf{lightweight computation} manner. The entire framework is trained via \textbf{end-to-end optimization} using a unified objective that balances task performance with sparsity, maximizing efficiency while maintaining high prediction accuracy.

\subsection{Problem Formulation}
\label{subsec:problem_formulation}

Given a sequence of dense spatio-temporal data over $T$ historical steps, $X \in \mathbb{R}^{T \times C \times H \times W}$ (e.g., image data with $C$, $H$ and $W$ as channels, height and width, respectively), the goal of spatio-temporal prediction is to learn a mapping $f$ with learnable network parameters $\theta$ that predicts a future sequence $\hat{Y}=f(X; \theta)$. The objective is to minimize the discrepancy between the prediction $\hat{Y}$ and the ground truth $Y$.

While traditional methods process the entire input $X$, our framework enhances efficiency by introducing learnable, binary masks. The first, a \textbf{Data Mask} $M_{\text{data}} \in \{0, 1\}^{H \times W}$, nullifies spatial locations deemed uninformative. The second, a \textbf{Model Mask} $M_{\text{weight}} \in \{0, 1\}^{K}$, disables $K$ prunable structural units within the model (e.g., convolutional filters).

Consequently, our core optimization problem is to find the optimal parameters $\theta$, data mask $M_{\text{data}}$, and model mask $M_{\text{weight}}$ that minimize the task loss $\mathcal{L}_{\text{task}}$, subject to predefined sparsity constraints $s_d$ (for data) and $s_w$ (for model weights):
\begin{equation}
\label{eq:objective}
\min_{\theta, M_{\text{data}}, M_{\text{weight}}} \quad \mathcal{L}_{\text{task}}(f(X \odot M_{\text{data}}; \theta \odot M_{\text{weight}}), Y)
\end{equation}
\begin{equation}
\label{eq:constraints}
\text{s.t.} \quad \frac{\|M_{\text{data}}\|_0}{H \times W} \le 1 - s_d, \quad \frac{\|M_{\text{weight}}\|_0}{K} \le 1 - s_w
\end{equation}
where $\odot$ denotes the mask application, and $\|\cdot\|_0$ is the L0 norm, which quantifies sparsity. This formulation defines our objective of trading off accuracy with data and model sparsity.

\subsection{Collaborative Pruning Mechanism}
\label{subsec:collaborative_pruning}

The innovation of \method{} lies in its collaborative pruning mechanism, which ensures the synergistic generation of data and model masks. This is achieved via an end-to-end trainable pipeline based on importance scores.

\subsubsection{Importance Score-based Mask Generation}
\label{ssubsec:importance_scores}

Directly optimizing discrete masks is computationally infeasible due to the combinatorial and non-differentiable nature of the problem. We circumvent this by introducing a continuous proxy variable: an \textbf{Importance Score Map} $S \in [0, 1]^{H \times W}$. This map is dynamically generated by a lightweight \textbf{Mask Generator}, $g_{\text{mask}}(\cdot; \phi)$, from the input $X$. Each element $S_{i,j}$ quantifies the importance of the spatial location $(i,j)$ for the final prediction.
\begin{equation}
\label{eq:score_map}
S = g_{\text{mask}}(X; \phi)
\end{equation}
This score map $S$ serves as the bridge connecting data and model pruning. The mask generator parameters $\phi$ are trained end-to-end with the entire network through backpropagation. This joint training ensures that the importance scores are learned specifically to benefit the main prediction task.

\subsubsection{Synergy and Synchronization}
\label{ssubsec:synergy}

Synergy is achieved by deriving both masks from the same importance score map $S$, ensuring their intrinsic consistency. The \textbf{Data Mask} is obtained from $S$; during training, we use $S$ as a soft mask ($X' = X \odot S$) for smooth gradient flow. For inference, we binarize $S$ into a hard mask $M_{\text{data}}$, a process simulated during training using the \textbf{Straight-Through Estimator (STE)} technique~\cite{yin2019understanding}. The \textbf{Model Mask} $M_{\text{weight}}$ is also tightly coupled with $S$, enabling input-aware structured pruning. We compute a ``weighted importance" scalar $I_k$ for each structural unit (e.g., $k$-th filter) by aggregating the scores in $S$ within its receptive field (e.g., via summation or averaging). Units with importance scores below a learnable threshold $\tau$ are pruned. This threshold $\tau$ is not a fixed hyperparameter but is treated as a learnable parameter, optimized jointly with the rest of the model via gradient descent. This mechanism ensures that units processing regions deemed unimportant by $S$ are likely to be pruned, thus synchronizing data and model pruning.

\subsection{End-to-End Optimization and Lightweight Inference}
\label{subsec:optimization}

We integrate all components of the \method{} framework through a unified loss function, enabling end-to-end optimization with standard gradient-based methods.

\subsubsection{Unified Loss Function}
\label{ssubsec:loss_function}

As shown in Figure~\ref{fig:framework}, a multi-objective loss function $\mathcal{L}_{\text{total}}$ drives the optimization, balancing accuracy with sparsity. It includes the \textbf{Task Loss} ($\mathcal{L}_{\text{task}}$) (e.g., MSE), the \textbf{Data Sparsity Loss} ($\mathcal{L}_{\text{sparsity}}(S)$), and the \textbf{Model Sparsity Loss} ($\mathcal{L}_{\text{sparsity}}(I)$). We use the L1 norm as a differentiable proxy to encourage sparsity in the importance score map $S$ and the unit importance scores $I$. The complete loss is:
\begin{equation}
\label{eq:total_loss}
\mathcal{L}_{\text{total}} = \mathcal{L}_{\text{task}} + \lambda_d \cdot \|S\|_1 + \lambda_w \cdot \|I\|_1
\end{equation}
where $\lambda_d$ and $\lambda_w$ are hyperparameters that control the trade-off between performance and efficiency.

\subsubsection{Training and Inference Pipeline}
\label{ssubsec:pipeline}

During the \textbf{Training Phase}, model parameters $\theta$ and mask generator parameters $\phi$ are optimized simultaneously. In each iteration, the mask generator produces a score map $S$ from the input data $X$, from which soft masks for data and model are derived. These masks are applied to yield a sparsified output $\hat{Y}$, and the unified loss is computed. The resulting gradients guide both the prediction model and the mask generator, as shown by the "End-to-End Optimization" loop.

During the \textbf{Inference Phase}, the pipeline is highly efficient. The trained mask generator performs a single forward pass to produce $S$, which is then binarized into hard masks $M_{\text{data}}$ and $M_{\text{weight}}$. These masks ensure that computation is performed only on important regions and units. This ``Lightweight Computation" process substantially reduces inference latency and memory footprint, making it ideal for resource-constrained edge scenarios.

We conduct extensive experiments to systematically answer three core research questions (RQs): \textbf{(RQ1) Effectiveness:} Does our method achieve a superior accuracy-efficiency trade-off compared to baselines? \textbf{(RQ2) Generality:} Can \method{} be seamlessly applied to diverse model architectures? \textbf{(RQ3) Mechanism Analysis:} Are the key designs within \method{}, particularly synergy, critical to its success?

\begin{table*}[t]
\centering
\small % 保持标准字号
\renewcommand{\arraystretch}{1.3} % 增加行高，提升阅读舒适度
\setlength{\tabcolsep}{8pt} % 优化列间距

\caption{Comprehensive comparison of \method{} against baselines. Lower is better for all metrics. 
Percentages in brackets indicate deltas relative to the \textbf{Dense} model (Red: error increase $\uparrow$; Blue: reduction $\downarrow$).}
\label{tab:main_results}

\begin{tabular}{ll ccc cc}
\toprule
\multirow{2}{*}{\textbf{Backbone}} & \multirow{2}{*}{\textbf{Method}} & \multicolumn{3}{c}{\textbf{Prediction Accuracy (MSE/MAE) $\downarrow$}} & \multicolumn{2}{c}{\textbf{Efficiency Metrics $\downarrow$}} \\
\cmidrule(lr){3-5} \cmidrule(lr){6-7}
& & \textbf{WeatherBench} & \textbf{SEVIR} & \textbf{TaxiBJ} & \textbf{GFLOPs} & \textbf{Latency (ms)} \\
\midrule

% --- SimVP 部分 ---
\multirow{3}{*}{\shortstack[l]{\textbf{SimVP} \\ (CNN)}} & Dense & 0.0452 & 24.51 & 2.51 & 120.4 & 85.2 \\
& MP~\cite{kumar2021pruning} & 0.0471 {\color{red}\scriptsize (+4.2\%)} & 25.89 {\color{red}\scriptsize (+5.6\%)} & 2.63 {\color{red}\scriptsize (+4.8\%)} & 31.5 {\color{blue}\scriptsize (-73.8\%)} & 42.1 {\color{blue}\scriptsize (-50.6\%)} \\
& \cellcolor[HTML]{EBF5FB}\textbf{\method{}} & \cellcolor[HTML]{EBF5FB}\textbf{0.0458} {\color{red}\scriptsize (+1.3\%)} & \cellcolor[HTML]{EBF5FB}\textbf{24.83} {\color{red}\scriptsize (+1.3\%)} & \cellcolor[HTML]{EBF5FB}\textbf{2.55} {\color{red}\scriptsize (+1.6\%)} & \cellcolor[HTML]{EBF5FB}\textbf{30.2} {\color{blue}\scriptsize (-74.9\%)} & \cellcolor[HTML]{EBF5FB}\textbf{33.8} {\color{blue}\scriptsize (-60.3\%)} \\
\addlinespace[4pt]

% --- ConvLSTM 部分 ---
\multirow{3}{*}{\shortstack[l]{\textbf{ConvLSTM} \\ (RNN)}} & Dense & 0.0515 & 28.14 & 2.85 & 155.8 & 110.5 \\
& MP~\cite{kumar2021pruning} & 0.0548 {\color{red}\scriptsize (+6.4\%)} & 30.02 {\color{red}\scriptsize (+6.7\%)} & 3.03 {\color{red}\scriptsize (+6.3\%)} & 40.1 {\color{blue}\scriptsize (-74.3\%)} & 58.3 {\color{blue}\scriptsize (-47.2\%)} \\
& \cellcolor[HTML]{EBF5FB}\textbf{\method{}} & \cellcolor[HTML]{EBF5FB}\textbf{0.0523} {\color{red}\scriptsize (+1.6\%)} & \cellcolor[HTML]{EBF5FB}\textbf{28.59} {\color{red}\scriptsize (+1.6\%)} & \cellcolor[HTML]{EBF5FB}\textbf{2.89} {\color{red}\scriptsize (+1.4\%)} & \cellcolor[HTML]{EBF5FB}\textbf{39.5} {\color{blue}\scriptsize (-74.6\%)} & \cellcolor[HTML]{EBF5FB}\textbf{41.7} {\color{blue}\scriptsize (-62.3\%)} \\
\addlinespace[4pt]

% --- TAU 部分 ---
\multirow{3}{*}{\shortstack[l]{\textbf{TAU} \\ (Transformer)}} & Dense & 0.0439 & 23.98 & 2.42 & 180.2 & 135.8 \\
& MP~\cite{kumar2021pruning} & 0.0465 {\color{red}\scriptsize (+5.9\%)} & 25.51 {\color{red}\scriptsize (+6.4\%)} & 2.58 {\color{red}\scriptsize (+6.6\%)} & 46.2 {\color{blue}\scriptsize (-74.4\%)} & 75.1 {\color{blue}\scriptsize (-44.7\%)} \\
& \cellcolor[HTML]{EBF5FB}\textbf{\method{}} & \cellcolor[HTML]{EBF5FB}\textbf{0.0447} {\color{red}\scriptsize (+1.8\%)} & \cellcolor[HTML]{EBF5FB}\textbf{24.37} {\color{red}\scriptsize (+1.6\%)} & \cellcolor[HTML]{EBF5FB}\textbf{2.46} {\color{red}\scriptsize (+1.7\%)} & \cellcolor[HTML]{EBF5FB}\textbf{45.5} {\color{blue}\scriptsize (-74.8\%)} & \cellcolor[HTML]{EBF5FB}\textbf{50.2} {\color{blue}\scriptsize (-63.0\%)} \\

\bottomrule
\end{tabular}
\end{table*}

\subsection{Experimental Setup}
\label{subsec:setup}

\noindent\textbf{\textit{Datasets}.} We evaluate on three public benchmarks with diverse spatio-temporal dynamics: \textbf{WeatherBench}~\cite{rasp2023weatherbench} for medium-range weather forecasting with large redundant areas, \textbf{SEVIR}~\cite{veillette2020sevir} for nowcasting of sparse and bursty weather events, and \textbf{TaxiBJ}~\cite{zhang2017deep} for fine-grained urban traffic prediction.

\noindent\textbf{\textit{Backbones and Baselines}.} To verify generality (RQ2), we integrate \method{} into three representative backbones: \textbf{SimVP}~\cite{gao2022simvp} (CNN-based), \textbf{ConvLSTM}~\cite{shi2015convolutional} (RNN-based), and \textbf{TAU}~\cite{tan2023temporal} (Transformer based). For a rigorous comparison (RQ1), we benchmark against two baselines for each backbone: the original \textbf{Dense Model} and a \textbf{Model Pruning (MP)} variant using L1-norm filter pruning~\cite{kumar2021pruning} at a comparable sparsity level.

\noindent\textbf{\textit{Metrics and Details}.} We evaluate \textbf{prediction accuracy} using Mean Squared Error (MSE) and Mean Absolute Error (MAE) (lower is better). For \textbf{computational efficiency}, we report theoretical GFLOPs and practical inference latency (ms) on an NVIDIA Jetson AGX Orin. All experiments are trained on 4 NVIDIA A100 GPUs using the AdamW optimizer. For \method{}, we set a target data sparsity $s_d = 0.7$ and model sparsity $s_w = 0.7$. Our code will be publicly available.

\subsection{Main Results and Analysis}
\label{subsec:main_results}

The comprehensive performance of \method{} and the baselines are summarized in Table~\ref{tab:main_results}.
\noindent\textbf{\method{} consistently achieves a superior accuracy-efficiency trade-off (RQ1).} The results clearly show that across all scenarios, \method{} significantly outperforms traditional model pruning (MP). At comparable GFLOPs reduction (around 75\%), \method{}'s accuracy drop is minimal, consistently staying within 1-2\% of the dense model. In contrast, MP often incurs a much larger accuracy degradation of 4-7\%. This demonstrates that our collaborative pruning mechanism more intelligently preserves model performance by removing systemic redundancy.

\noindent\textbf{\method{} exhibits strong generality across diverse architectures (RQ2).} Our framework successfully enhances the efficiency of CNN, RNN, and Transformer-based models with a consistent pattern of improvement. This confirms that \method{} is a universal, model-agnostic efficiency framework, not a specialized optimization for a particular architecture. Furthermore, its practical speedup is particularly noteworthy; for instance, on the TAU backbone, \method{} reduces latency by 63.0\% from the dense model, a far greater improvement than MP's 44.7\%, despite similar GFLOPs. This is because co-pruning reduces both computation and data access overhead, a crucial advantage on real hardware.

\subsection{Robustness and Reliability Analysis}
To address concerns regarding the reliability of our dynamic mask generator, we evaluate its performance under environmental noise and intentional mask failures. As illustrated in Table~\ref{tab:robustness_wide}, \method{} exhibits remarkable noise resilience. When subjected to heavy Gaussian noise ($\sigma=0.05$), the MSE of \method{} increases by only 6.7\%, compared to 13.3\% for the dense model. This suggests that the learned importance map $S$ effectively filters out high-frequency noise in non-critical regions. 

Furthermore, we simulate ``mask failures'' by intentionally flipping pixels from ``important'' to ``pruned''. Even with a 10\% failure rate in critical dynamic regions, \method{} maintains superior performance over the random pruning baseline, demonstrating the inherent fault tolerance captured by our collaborative pruning mechanism.

\begin{table*}[t] % 跨双栏
\centering
\renewcommand{\arraystretch}{1.3}
\setlength{\tabcolsep}{18pt} % 增加列间距，使通栏表格不显得空旷

\caption{Robustness analysis against environmental noise and internal mask failures. \textbf{Bold} indicates the best performance under each condition.}
\label{tab:robustness_wide}

\begin{tabular}{l c cc cc}
\toprule
\multirow{2}{*}{\textbf{Method}} & \textbf{Clean Baseline} & \multicolumn{2}{c}{\textbf{Gaussian Noise ($\sigma$)}} & \multicolumn{2}{c}{\textbf{Mask Failure Rate}} \\
\cmidrule(lr){2-2} \cmidrule(lr){3-4} \cmidrule(lr){5-6}
 & (MSE $\downarrow$) & 0.01 & 0.05 & 5\% & 10\% \\
\midrule
Dense Model & 0.0452 & 0.0468 {\color{gray}\scriptsize (+3.5\%)} & 0.0512 {\color{gray}\scriptsize (+13.3\%)} & -- & -- \\
Random Pruning (70\%) & 0.0471 & 0.0495 {\color{gray}\scriptsize (+5.1\%)} & 0.0584 {\color{gray}\scriptsize (+24.0\%)} & 0.0510 & 0.0562 \\
\rowcolor[HTML]{EBF5FB} 
\textbf{Dyna-Pruner (Ours)} & \textbf{0.0458} & \textbf{0.0465} \scriptsize{\textbf{(+1.5\%)}} & \textbf{0.0489} \scriptsize{\textbf{(+6.7\%)}} & \textbf{0.0472} & \textbf{0.0491} \\
\bottomrule
\end{tabular}
\end{table*}

\begin{figure}[ht]
\centering
\includegraphics[width=\linewidth]{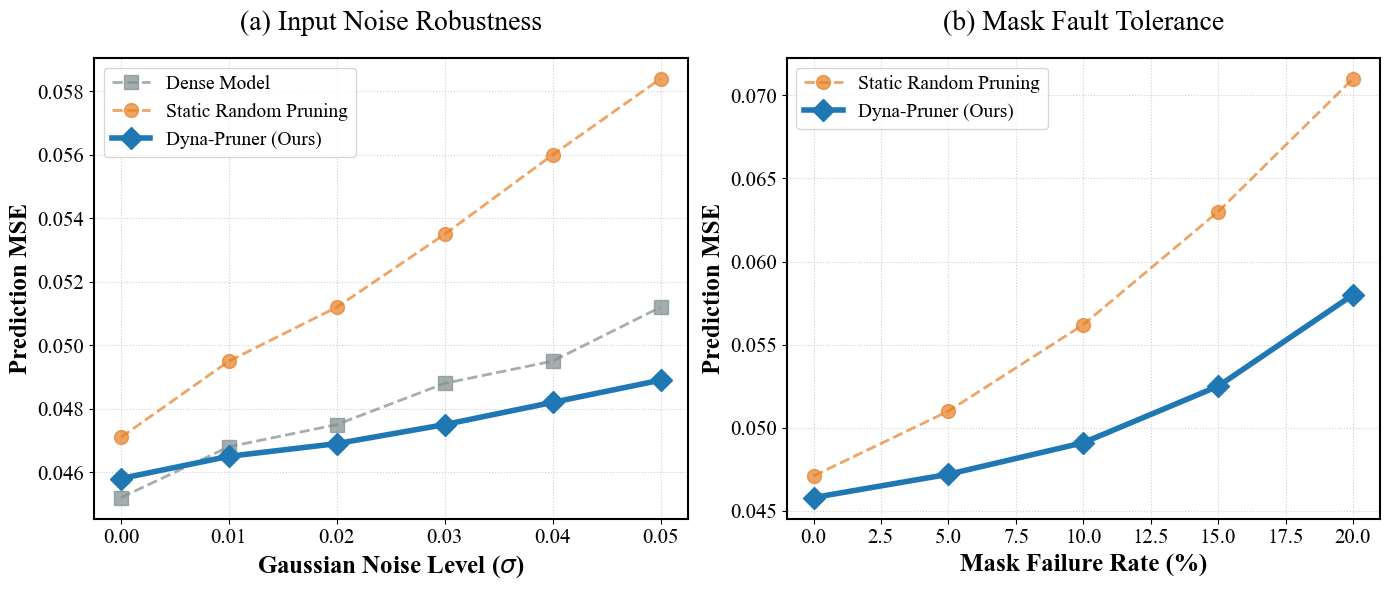}
\caption{Visualization of (a) robustness to increasing Gaussian noise and (b) performance degradation under simulated mask generator failures.}
\label{fig:robustness_viz}
\end{figure}

\subsection{Parameter Sensitivity Analysis}
\label{subsec:sensitivity}
\begin{table}[t] % 单栏环境
\centering
\tiny
\setlength{\tabcolsep}{3.5pt} % 显著缩小列间距以适应单栏
\renewcommand{\arraystretch}{1.2} % 保持适度的行高

\caption{Sensitivity analysis of sparsity targets on \textbf{WeatherBench and SEVIR} (SimVP backbone). Accuracy in MSE. Default setting is \textbf{bolded} and \color{blue}highlighted\color{black}.}
\label{tab:sensitivity}

\resizebox{\columnwidth}{!}{
\begin{tabular}{cc cc cc} % 去掉了垂直线 |，更符合顶会审美
\toprule
\multicolumn{2}{c}{\textbf{Sparsity Targets}} & \multicolumn{2}{c}{\textbf{Accuracy (MSE) $\downarrow$}} & \multicolumn{2}{c}{\textbf{Efficiency $\downarrow$}} \\
\cmidrule(lr){1-2} \cmidrule(lr){3-4} \cmidrule(lr){5-6}
\textbf{$s_d$} & \textbf{$s_w$} & \textbf{Weather} & \textbf{SEVIR} & \textbf{GFLOPs} & \textbf{Latency} \\
\midrule
0.5 & 0.5 & 0.0454 & 24.75 & 58.1 & 55.3 \\
0.7 & 0.5 & 0.0456 & 24.80 & 45.9 & 48.1 \\

\rowcolor[HTML]{EBF5FB} 
\textbf{0.7} & \textbf{0.7} & \textbf{0.0458} & \textbf{24.83} & \textbf{30.2} & \textbf{33.8} \\

0.9 & 0.7 & 0.0469 & 25.15 & 21.5 & 28.4 \\
0.9 & 0.9 & 0.0482 & 25.62 & 12.3 & 21.7 \\
\bottomrule
\end{tabular}
}
\end{table}
\noindent\textbf{Impact of Sparsity Targets.} To investigate the impact of the core hyperparameters in our framework, we conduct a sensitivity analysis on the data sparsity target ($s_d$) and model sparsity target ($s_w$). We vary these targets and evaluate the performance on two representative datasets, WeatherBench and SEVIR, using the SimVP backbone. As shown in Table~\ref{tab:sensitivity}, the results demonstrate a clear and consistent trade-off between efficiency and accuracy across both datasets. Increasing the sparsity targets from 0.5 to 0.9 progressively reduces GFLOPs and latency, but also leads to a gradual increase in prediction error (MSE). Our chosen default setting ($s_d=0.7, s_w=0.7$) strikes a strong balance, achieving over a 75\% reduction in GFLOPs while maintaining a minimal accuracy loss. This consistency across different data distributions indicates that Dyna-Pruner is robust within a reasonable range of sparsity levels, and these targets provide an intuitive mechanism for users to customize the model's performance based on specific hardware constraints and accuracy requirements.

\subsection{Ablation and Visualization Analysis}
\label{subsec:ablation_viz}

To answer \textbf{RQ3 (Mechanism Analysis)}, we ablate key components of \method{} and visualize its learned behavior.

\noindent\textbf{The synergy mechanism is critical for performance.} As shown in Table~\ref{tab:ablation}, isolating data pruning (DP) or model pruning (MP) leads to a noticeable accuracy drop. More importantly, a naive, non-synergistic combination of DP and MP performs even worse, suggesting that uncoordinated pruning can remove complementary information. In contrast, our full framework with the synergy mechanism achieves the best of both worlds: the highest efficiency gains with the lowest accuracy loss, performing nearly on par with the dense model. This provides strong evidence that the synergy between data and model pruning is the core driver of \method{}'s effectiveness.

\begin{table}[ht]
\centering
\small % 顶会标准字号
\setlength{\tabcolsep}{5pt} % 优化列间距
\renewcommand{\arraystretch}{1.3} % 增加行高
\caption{Ablation study of the collaborative pruning mechanism on WeatherBench. ``Independent'' refers to a naive combination of DP and MP without shared synchronization.}
\label{tab:ablation}

\begin{tabular}{ccc ccc}
\toprule
\multicolumn{3}{c}{\textbf{Components}} & \multicolumn{3}{c}{\textbf{Metrics}} \\
\cmidrule(lr){1-3} \cmidrule(lr){4-6}
\textbf{DP} & \textbf{MP} & \textbf{Synergy} & \textbf{MSE $\downarrow$} & \textbf{GFLOPs $\downarrow$} & \textbf{Latency $\downarrow$} \\
\midrule
\xmark & \xmark & -- & 0.0452 & 120.4 & 85.2 \\
\cmark & \xmark & -- & 0.0465 & 65.1 & 58.7 \\
\xmark & \cmark & -- & 0.0471 & 31.5 & 42.1 \\
\midrule
\cmark & \cmark & \xmark & 0.0469 & 30.8 & 35.5 \\
\rowcolor[HTML]{EBF5FB} 
\cmark & \cmark & \cmark & \textbf{0.0458} & \textbf{30.2} & \textbf{33.8} \\
\bottomrule
\end{tabular}
\end{table}

\begin{figure}[htbp]
    \centering
    \includegraphics[width=\columnwidth]{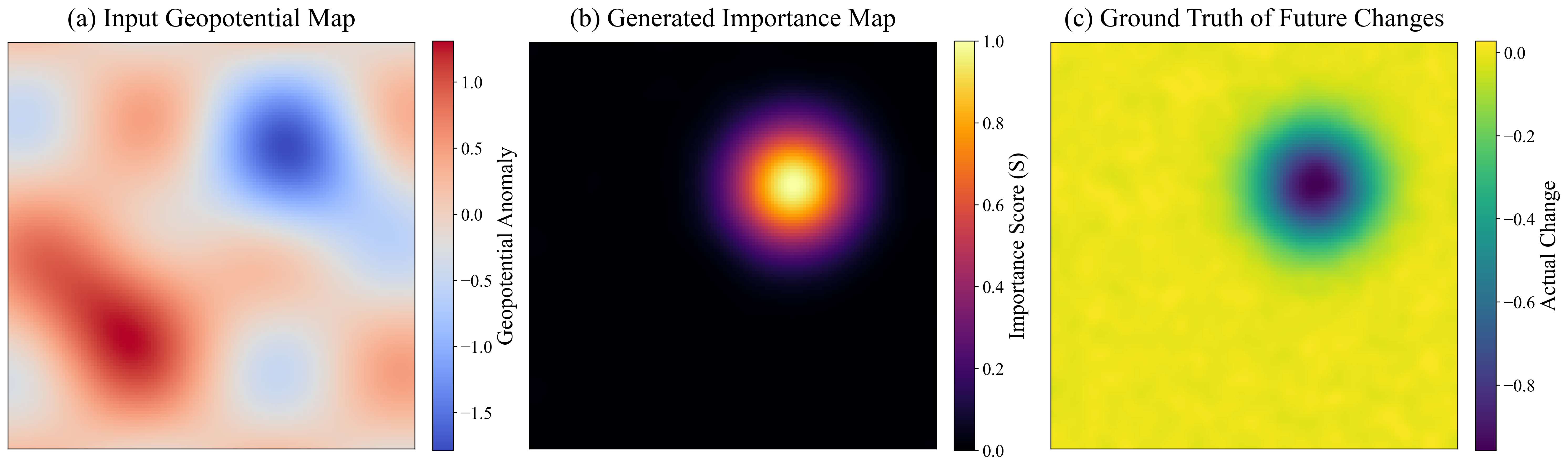}
    \caption{Importance visualization on \textbf{WeatherBench}.
\textbf{(a)} Input geopotential map. \textbf{(b)} Importance score map $S$ (brighter = higher importance).
\textbf{(c)} Magnitude of future change $|\Delta|$ (brightness shows change intensity; warm/cool colors indicate positive/negative values).
Bright regions in (b) spatially align with high-magnitude changes in (c), demonstrating that \method{} allocates computational resources to meteorologically active regions.}
    \label{fig:visualization}
\end{figure}

\noindent\textbf{\method{} learns to focus on physically meaningful dynamic regions.} The visualization in Figure~\ref{fig:visualization} provides intuitive insight into our method. The generated importance map (b) clearly highlights regions corresponding to active weather systems (e.g., cyclones) in the input (a), which are precisely the areas where future changes are most significant (c). Simultaneously, \method{} learns to ignore vast, static background areas, such as calm oceans. This alignment with meteorological priors demonstrates that our framework learns a physically interpretable and effective attention mechanism for dynamic resource allocation.

\section{Conclusion}
\label{sec:conclusion}

In this paper, we introduced \method{}, a collaborative pruning framework that addresses the computational overhead and data redundancy in spatio-temporal prediction. By establishing an \textbf{input-dependent synchronization} between data sparsity and structural compression, \method{} dynamically adapts the model's computational graph to the information density of each specific sample. Evaluated across {CNN, RNN, and Transformer} backbones, our framework consistently delivers up to a {70\% FLOPs reduction} and a {2.5$\times$ practical speedup} on edge hardware with negligible accuracy loss ($<$1\%). The success of \method{} underscores that the {synergy between data and model streamlining} is more effective than isolated compression techniques. Ultimately, our work provides a practical paradigm for deploying high-fidelity predictive models in resource-constrained environments, such as IoT sensors and mobile edge devices.

% \clearpage
\bibliographystyle{IEEEbib}
\bibliography{references}

\end{document}